\documentclass[letterpaper, 10 pt, conference]{ieeeconf}  

\IEEEoverridecommandlockouts                              
\usepackage[pdftex]{graphicx} 

\usepackage{amsthm}
\usepackage{amsmath} 
\usepackage{amssymb}  
\usepackage{bm}  
\usepackage{color}
\usepackage{xcolor}
\usepackage{siunitx}
\PassOptionsToPackage{hyphens}{url}\usepackage[hidelinks]{hyperref}
\usepackage{comment}
\usepackage{physics}

\newcommand{\rt}{\textcolor{red}}

\begin{document}

\title{\LARGE \bf RL-augmented MPC Framework for Agile and Robust Bipedal Footstep Locomotion Planning and Control}
\author{Seung Hyeon Bang$^{*1}$, Carlos Arribalzaga Jové$^{*1,2}$, and Luis Sentis$^{1}$ 
\thanks{$^{*}$ These authors have contributed equally to this work.} 
\thanks{$^{1}$ The University of Texas at Austin, TX, USA, 
\tt\small \{bangsh0718, ca36828, lsentis\}@utexas.edu}
\thanks{$^{2}$ Universitat Politècnica de Catalunya (UPC), Barcelona, Spain}}%

\maketitle

\thispagestyle{empty}
\pagestyle{empty}



\begin{abstract}
    This paper proposes an online bipedal footstep planning strategy that combines model predictive control (MPC) and reinforcement learning (RL) to achieve agile and robust bipedal maneuvers. While MPC-based foot placement controllers have demonstrated their effectiveness in achieving dynamic locomotion, their performance is often limited by the use of simplified models and assumptions. To address this challenge, we develop a novel foot placement controller that leverages a learned policy to bridge the gap between the use of a simplified model and the more complex full-order robot system. Specifically, our approach employs a unique combination of an ALIP-based MPC foot placement controller for sub-optimal footstep planning and the learned policy for refining footstep adjustments, enabling the resulting footstep policy to capture the robot's whole-body dynamics effectively. This integration synergizes the predictive capability of MPC with the flexibility and adaptability of RL. We validate the effectiveness of our framework through a series of experiments using the full-body humanoid robot DRACO 3. The results demonstrate significant improvements in dynamic locomotion performance, including better tracking of a wide range of walking speeds, enabling reliable turning and traversing challenging terrains while preserving the robustness and stability of the walking gaits compared to the baseline ALIP-based MPC approach.
    
\end{abstract}

\section{INTRODUCTION}
\label{sec:introduction}
Agile and robust bipedal locomotion is essential for humanoid robots to achieve human-level performance. One of the main challenges in achieving this is designing a footstep policy that enables bipeds to constantly adjust their planned footstep positions to maintain balance as well as to achieve more agile and fast maneuvers, even while traversing adverse environments, such as external disturbances or challenging terrains.

\begin{figure}[t!]
    \centering
    \includegraphics[width=\columnwidth]{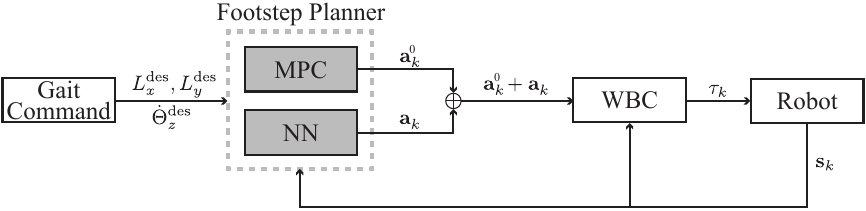}
    \caption{\textbf{Overview of the proposed control framework:} The footstep planner consists of a simplified model-based model predictive controller (MPC) and a neural network (NN). Together, these components generate the footstep policy by integrating the solutions from each module. A whole-body feedback controller (WBC) then tracks the footstep policy.}
    \label{fig:framework_overview}
\end{figure}

In this paper, we present an RL-augmented MPC framework designed to generate a footstep policy for agile and robust bipedal locomotion. Our framework, as illustrated in Fig.~\ref{fig:framework_overview}, combines model-based optimal control (MBOC) with reinforcement learning (RL) to leverage the strengths of both approaches. Specifically, we adopt a hierarchical control architecture consisting of a high-level (HL) planner, which integrates both MPC and RL policies, and a low-level (LL) tracking controller. The MPC utilizes a simplified model to generate an initial, suboptimal footstep plan. The residual RL policy then refines this plan by leveraging the robot's full-order dynamics model. This approach helps overcome the modelling errors of the simplified model and thus results in an enhanced footstep policy.

\subsection{Related Work:}
In legged robot locomotion, footstep planning is essential for maintaining stability in the robot's inherently unstable dynamics while navigating through complex environments. The Raibert heuristic~\cite{Raibert1983StableLocomotion} has been widely used in legged robot control due to its simplicity and effectiveness. 

Alternatively, model-based footstep planning strategies have increasingly been developed. These methods typically use simplified models that approximate the robot's underactuated dynamics to determine foot placements that stabilize the robot's CoM dynamics.~\cite{Ahn2019ControlActuators, Kim2020DynamicControl} utilized the Linear Inverted Pendulum (LIP) model to calculate footstep placement based on the robot's CoM states, aiming to reverse the CoM velocity after a specified duration. On the other hand, the capture point concept \cite{Pratt2006CaptureRecovery, Koolen2012Capturability-basedModels}, based on the divergent component of motion of the LIP model, has been widely used to develop footstep decision-making strategies. It provides a theoretical framework for analyzing a controller's ability to bring robots to a stop within a specified number of steps.

On the other hand,~\cite{Powell2016Mechanics-basedConstraints} uses the Angular Momentum Linear Inverted Pendulum (ALIP) model to generate a gait for stable walking. This gait is designed by generating the robot’s swing foot and vertical CoM trajectories while conserving angular momentum at impact.~\cite{Gong2022ZeroLocomotion} proposes a one-step ahead gait controller that determines where to step given a desired angular momentum.~\cite{Gibson2022Terrain-AdaptiveConstraints} extended this gait controller with an MPC-based framework~\cite{Lee2020MPC-BasedTasks, Lee2023Real-TimeControl, Bang2024VariableManeuvers}, considering workspace and friction cone constraints. More recently, more complex dynamic models such as the single-rigid body model~\cite{Bledt2020ExtractingControl} and the centroidal dynamics model~\cite{Romualdi2022OnlineAdjustment} have been adopted to jointly optimize the states and footsteps. However, the abstractions and assumptions in these simplified models can compromise the robustness of the optimized footsteps or require heuristics and arduous tuning due to the lack of consideration of the full-order model. 

In contrast to model-based approaches, model-free RL-based approaches can discover a footstep policy through trial and error, eliminating the need for explicit dynamic models. RL enables the discovery of the policy by maximizing the discounted cumulative reward as an RL agent interacts with the environment. Unlike end-to-end RL approaches~\cite{Xie2018FeedbackLearning,Siekmann2020LearningLocomotion, Li2021ReinforcementRobots} which learn locomotion policies by directly mapping sensor data to joint commands, RL-based footstep policies are usually developed within a hierarchical control structure~\cite{Peng2017DeepLoco:Learning, Green2021LearningModel, Duan2021LearningLocomotion} that incorporates a HL and a LL controller. For example,~\cite{Peng2017DeepLoco:Learning} trains a HL controller to generate footsteps using proprioceptive states and terrain information as input. Similarly,~\cite{Duan2021LearningLocomotion} proposes an RL method to train a residual footstep policy with an offline gait library. These approaches have enabled the direct exploration of task space actions, yielding more sample-efficient learning. Following this paradigm,~\cite{Castillo2023TemplateLocomotion} formulates observation and action spaces of a Markov Decision Process (MDP) inspired by the ALIP model to train a footstep policy that enables robust walking under adversarial conditions. However, unlike model-based footstep planning, model-free RL approaches do not have access to the underlying robot dynamics and so struggle with sample inefficiency as they must learn to reason about the robot dynamics implicitly. As a result, they require substantial training data since they need to build the policies from scratch.

In recent years, there has been extensive research on integrating MBOC and RL to leverage the benefits from both approaches.~\cite{Kang2023RLLocomotion} proposed an RL framework that incorporated MBOC to generate both reference base and foot motions and employed the motion imitation technique to learn complex locomotion tasks.~\cite{Chen2023ReinforcementRobots} utilized RL to learn task-optimal parameters of a simplified dynamics model, which was subsequently optimized to generate CoM trajectories and footstep positions using an MPC framework. 
In~\cite{Ahn2020Data-EfficientModel}, an RL footstep policy was trained with guidance from the previously-mentioned Time-to-Velocity Reversal (TVR) footstep planner, based on the LIP model. The training process utilized the TVR planner's solution as suboptimal guidance, enabling the residual RL policy to maximize long-term rewards without requiring excessive training data.

Our approach combines MBOC and RL, similarly to the method introduced in~\cite{Ahn2020Data-EfficientModel}. However, instead of adopting the LIP model, we use an ALIP model-based MPC footstep generator similar to~\cite{Gibson2022Terrain-AdaptiveConstraints}. This approach allows us to obtain more viable footsteps to feed to the RL policy, thanks to the MPC's handling of strict constraints and its use of a prediction horizon. Unlike~\cite{Ahn2020Data-EfficientModel} computing the next desired footstep at the apex of each step, our method enables replanning footsteps multiple times during the foot swing motion. This capability aims to be more versatile for robust locomotion behaviors under external disturbances. Moreover, in contrast to~\cite{Castillo2023TemplateLocomotion}, our RL policy leveraging the MBOC process converges faster during training, enabling sample-efficient learning. 
\vspace{-0.09in}        
\subsection{Contributions:} The main contributions of this paper are the following:
\begin{enumerate}
    \item We propose the first RL-augmented MPC framework for bipedal footstep generation, designed to significantly enhance the tracking of the robot's walking speed, the robustness to external disturbances, the walking adaptability (transitioning between different velocity commands), and the ability to traverse arbitrary slopes.
    \item We designed flexible reward terms for the RL process to effectively learn from the ALIP-MPC process.
    \item We demonstrate that our approach achieves more agile, robust, and adaptive locomotion behaviors than using MPC alone by overcoming modelling errors associated with the simplified dynamics used by the MPC footstep planner. 
\end{enumerate}
\vspace{-0.10in}
\subsection{Organization:} The remainder of this paper is organized as
follows. Section~\ref{sec:preliminaries} provides a concise overview of the ALIP-based MPC footstep planner, WBC, and model-free RL. Section~\ref{sec:problem_statement} describes the locomotion problem addressed in this study. In Section~\ref{sec:solution_approach}, we describe the proposed RL-augmented MPC approach. Section~\ref{sec:experimental_results} validates the effectiveness of the proposed approach through various locomotion scenarios. Finally, Section~\ref{sec:conclusion} concludes the paper and discusses future works.

\section{PRELIMINARIES}
\label{sec:preliminaries}
In this section, we introduce the ALIP-based MPC footstep planner~\cite{Gibson2022Terrain-AdaptiveConstraints}, which we will use to synergize with our RL method. We also  present background of the whole-body controller (WBC) and model-free RL approaches that we employ.

\subsection{ALIP-based MPC Footstep Planner} \label{sec:alip_mpc}
The main objective of the ALIP-based MPC footstep planner is to compute the desired footholds to enable the angular momentum to converge to a desired state at the end of each step. In the following subsection, we briefly describe the main components of the MPC.

\subsubsection{ALIP Model} \label{sec:alip_model}
Simplified dynamic models offer the advantage of capturing the centroidal states of the robot while reducing the dimensions of the full-order model under several assumptions. Among the models used for footstep design, the ALIP model~\cite{Powell2016Mechanics-basedConstraints} has increasingly gained attention due to its higher accuracy for state estimations of the model. 
This advantage is attributed to the use of the robot's angular momentum about the contact point, which is represented as one of its states in the model. Under the model  assumptions, with one of them being that the angular momentum about the center of mass ($\bm{L}^c$) is zero to facilitate linearization~\cite{Gibson2022Terrain-AdaptiveConstraints}, the ALIP dynamics model is given by:
\begin{align}\label{eq:alip_model}
    \mathbf{\dot{x}} = \mathbf{A} \mathbf{x} + \mathbf{B} \mathbf{u}_{\textrm{fp}}
\end{align}
where
\begin{equation*}
    \begin{aligned}
        \mathbf{A} = \begin{bmatrix}
                     0 & 0 & 0 & \frac{1}{mz_H} \\
                     0 & 0 & -\frac{1}{mz_H} & 0 \\
                     0 & -mg & 0 & 0 \\
                     mg & 0 & 0 & 0
                     \end{bmatrix},
        \mathbf{B} = \begin{bmatrix}
                     -1 & 0 \\
                     0 & -1 \\
                     0 & 0 \\
                     0 & 0 
                     \end{bmatrix},
    \end{aligned}
\end{equation*}
$m$ denotes the mass of the robot, $g$ is the gravitational constant, and $z_H$ is the constant CoM height. The state and input variables are defined as $\mathbf{x} = [x_c, y_c, L_x, L_y]^{\top} \in \mathbb{R}^4$ and $\mathbf{u}_{\textrm{fp}} = [u_x, u_y]^{\top} \in \mathbb{R}^2$, where $x_c, y_c$ represents the CoM position in the $x, y$-direction with respect to the stance contact point, $L_x, L_y$ denotes the angular momentum about the $x, y$-axes of the contact point, and $u_x, u_y$ are the new foot placement positions in the $x, y$-direction from the current stance contact point.

\subsubsection{MPC Formulation for Foot Placement}  

Assuming conservation of angular momentum about the contact point before and after each footstep, the resulting discrete-time dynamics over an $N_s$ footstep horizon, with each step having a fixed time duration of $T_s$, including the step-to-step and intra-step dynamics, are given as follows:
\begin{equation} \label{eq:discrete_dyn}
\mathbf{x}_{k+1} = 
\begin{cases}
    \begin{aligned}
        &e^{\mathbf{A}\Delta t}(\mathbf{x}_{k}+ \mathbf{B} \mathbf{u}_{\textrm{fp},k}), &\text{if $k = iN_{\Delta t}$}, \\
        &                            &\text{$0 \leq i \leq N_{s}-1$} \\
        &e^{\mathbf{A}\Delta t}\mathbf{x}_{k}, &\text{otherwise} 
    \end{aligned}
\end{cases}
\end{equation} \\
where $\Delta t$ is the sampling time and $N_{\Delta t} = T_s / \Delta t \in \mathbb{Z}$ is the number of samples for each step. 




The MPC process aims to find the desired footholds with the objective of tracking the desired angular momentum ($L_{x}^{\textrm{des}}, L_{y}^{\textrm{des}}$) during step transitions while satisfying kinematic constraints (e.g., leg length), $\mathcal{X}^{\textrm{kin}}$, friction cone constraints, $\mathcal{X}^{\textrm{slip}}$, and foot placement safety constraints, $\mathcal{U}$~\cite{Gibson2022Terrain-AdaptiveConstraints}. The MPC problem is formulated over $N_s$ footstep as follows:
\begin{equation} \label{eq:alip_mpc}
\begin{aligned}
    \min_{\mathbf{U}_{\text{fp}}} & \quad \lVert \mathbf{x}_{e,N_{\Delta t}N_{s}} \rVert^{2}_{\mathbf{Q}_{f}} + \sum^{N_{\Delta t}N_{s}-1}_{k = 0} \lVert \mathbf{x}_{e,k} \rVert ^{2}_{\mathbf{Q}_{k}} \\
    \textrm{subject to} 
    & \quad \eqref{eq:discrete_dyn}, \\ 
    & \quad \forall \mathbf{x}_{k} \in \mathcal{X}^{\textrm{kin}} \cup \mathcal{X}^{\textrm{slip}} \text{ and } \forall \mathbf{u}_{\textrm{fp},k} \in \mathcal{U},\\
    & \quad \mathbf{Q}_{k} = \mathbf{0}, \hspace{0.5em} \forall k \notin \{N_{\Delta t}, \dots, N_{\Delta t}\left(N_{s}-1\right)\} \\
    & \quad \mathbf{x}_0 = e^{\mathbf{A}T_r} R_{\textrm{MPC}}^\top \mathbf{x}_{\textrm{cm}}
\end{aligned}
\end{equation}
where $\mathbf{U}_{\textrm{fp}} = [\mathbf{u}_{\textrm{fp},0}^{\top}, \mathbf{u}_{\textrm{fp},N_{\Delta t}}^{\top},\cdots, \mathbf{u}_{\textrm{fp},N_{\Delta t}\left(N_s-1\right)}^{\top} ]^{\top}$ and $\mathbf{x}_{e,k}:=\mathbf{x}_{k} - \mathbf{x}_{k}^{\text{des}}$. The desired state at each step transition is constructed via the solution of a periodic orbit, similar to~\cite{Gong2022ZeroLocomotion,Gibson2022Terrain-AdaptiveConstraints}, where $L_x^{\textrm{des}}:=L_x^{\textrm{main}} + L_x^{\textrm{offset}}$, with $L_x^{\textrm{offset}}$ being an additional lateral angular momentum term. $\mathbf{x}_0$ is the predicted state just before impact during the current step using~\eqref{eq:alip_model}. $T_r$ denotes the remaining duration of the current step, $R_{\textrm{MPC}}$ is the rotation matrix considering the desired yaw orientation of the torso, and $\mathbf{x}_{\textrm{cm}}$ represents the current state measurement. We also note that the above MPC struggles with continuous turning motions within its $N_s$ steps prediction horizon because the ALIP model does not inherently consider the rotational dynamics~\cite{Orin2013CentroidalRobot}. Therefore, we utilize the torso yaw rate command $\dot{\Theta}_z^{\textrm{des}}$ to compute the desired foot orientation command $\gamma_k^0$ as a one-step ahead prediction, similarly to~\cite{Gong2022ZeroLocomotion}.
This MPC formulation is rewritten in the form of a quadratic program (QP). After solving this QP, only the first optimal solution, $\mathbf{u}_{\textrm{fp},0}$, is utilized. 

\subsection{Whole-body Control (WBC)} \label{sec: WBC}
Several approaches \cite{Sreenath2011AMABEL, Sentis2010CompliantRobots} exist for computing joint commands to execute high-level objectives (e.g., desired foot positions) in task space. In particular, given multiple task objectives, WBC \cite{Kim2020DynamicControl, Lee2022OnlineDisturbances, Bang2023ControlBody} takes into account the full-order dynamics model of a robot to compute the optimal joint commands that minimize the tracking error of the tasks hierarchically based on state feedback. It also considers several constraints, including contact constraints and actuator limits, and is usually formulated as inverse dynamics with QP to offer real-time computation.  

\subsection{Model-free Reinforcement Learning (RL)}
The RL problem is described as a Markov Decision Process (MDP) $\mathcal{M} = \left(\mathcal{S}, \mathcal{A}, P, R, \gamma \right)$, which consists of a state space~$\mathcal{S}$, an action space~$\mathcal{A}$, a state transition function $P: \mathcal{S} \times \mathcal{A} \times \mathcal{S} \rightarrow [0, 1]$, a reward function $R: \mathcal{S} \times \mathcal{A} \rightarrow \mathbb{R}$, and a discount factor~$\gamma$. The goal of RL is to find a policy parameterized by a parameter~$\theta$, $\pi^{\star}_{\theta}: \mathcal{S} \rightarrow \mathcal{A}$, that maximizes the expected discounted reward:
\begin{align}
    \pi^{\star}_{\theta} = \underset{\pi_{\theta}}{\mathrm{argmax}} \hspace{0.5em} \mathbb{E}_{\tau \sim \pi_{\theta}} \left[\sum_{k=0}^{\infty} \gamma^{k} R(s_k, a_k) \right],
\end{align}
where $s_k \in \mathcal{S}$, $a_k \in \mathcal{A}$, and $\tau \sim \pi_{\theta}$ is a trajectory drawn from the policy $\pi_{\theta}$.  

\section{PROBLEM STATEMENT}
\label{sec:problem_statement}

\begin{figure}[t!]
    \centering
    \includegraphics[width=\columnwidth]{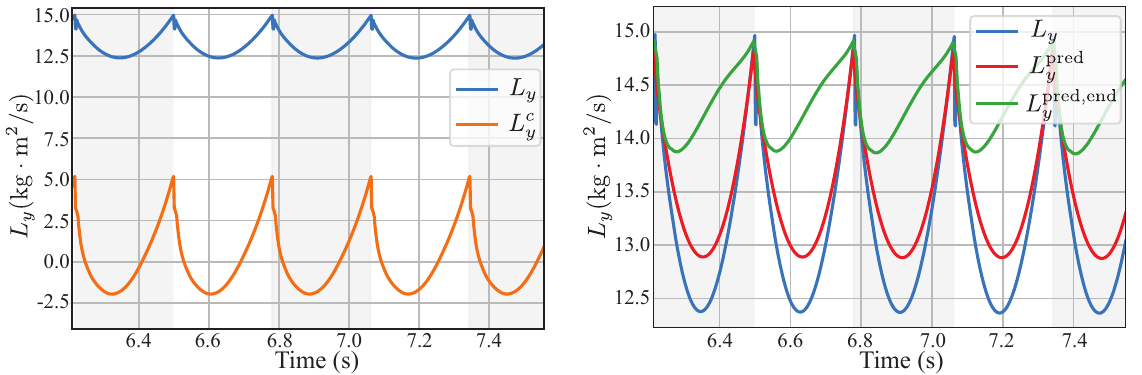}
    \caption{\textbf{ALIP-model limitation:} Comparison of the angular momentum about the robot's contact point, $L_y$, angular momentum about its CoM ($L_y^c$) (left), and the predicted evolution $L_y^{\textrm{pred}}$, using the forward simulation of the ALIP model. The initial state of $L_y$ is taken at the time of the step transition, and the predicted evolution at the end of the step, $L_y^{\textrm{pred, end}}$, is shown at every time instance during foot swing (right), while the robot walks forward using ALIP-based MPC~\cite{Gibson2022Terrain-AdaptiveConstraints} and WBC~\cite{Bang2023ControlBody}. Notice that the blue and red lines are noticeably different and $L_y^{\textrm{pred, end}}$ fluctuates considerably.}
    \label{fig:alip_limitation}
\end{figure}
\begin{figure*}[t!]
    \centering
    \includegraphics[width=\textwidth]{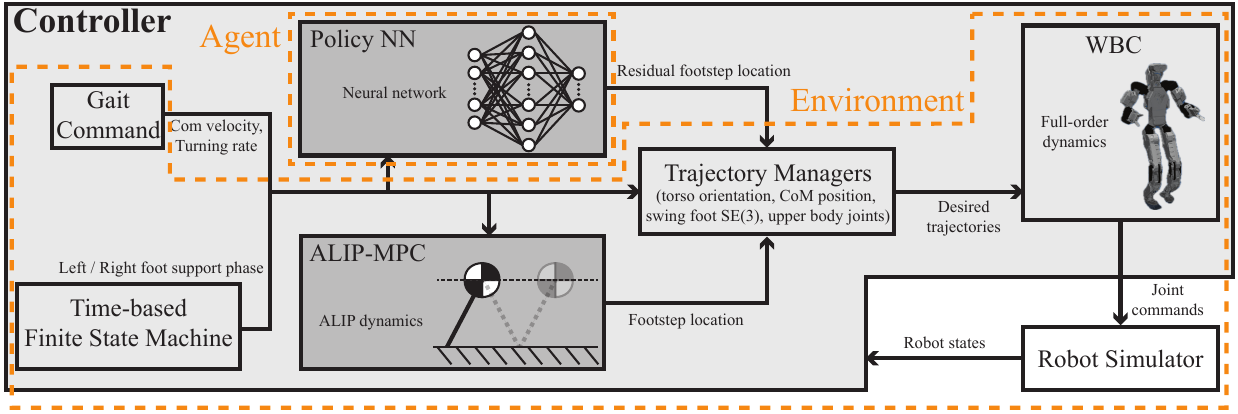}
    \caption{\textbf{Reinforcement learning framework:} The agent learns a policy neural network (NN) in simulation, where the robot is controlled through the modules in the controller. The ALIP-based model predictive controller (MPC) follows a gait command and operates based on a time-based finite state machine (FSM), which manages the footstep timing. The ALIP-MPC process optimizes the desired footstep location, while the trajectory managers generate the swing foot trajectories based on the footstep locations. The trajectory managers also govern the desired task trajectories to harness the full-order model. All desired trajectories are sent to the whole-body controller (WBC)~\cite{Bang2023ControlBody} to generate joint commands. In this learning framework, the policy NN generating residual footstep locations is the agent, while everything in the closed-loop system is considered the environment.} 
    \label{fig:rl_framework}
\end{figure*}
The effectiveness of the ALIP-based MPC process to generate an optimal foot location for agile and robust bipedal locomotion highly depends on the accuracy of the ALIP model relative to the full-order robot model. Although the ALIP model is effective for controlling a variety of bipedal robots~\cite{Gong2022ZeroLocomotion}, the assumptions underlying the model described in Section~\ref{sec:alip_model} are invalid for realistic robots, like our DRACO 3 robot, which has a relatively large distal mass on the leg~\cite{Bang2023ControlBody}, as illustrated in Fig.~\ref{fig:alip_limitation}. This model discrepancy results in a significant difference between the actual ($L_y$ simulated with the full-order model) and predicted ($L_y^{\textrm{pred}}$ simulated with the ALIP model) values, causing the predicted value at the end of the step ($L_y^{\textrm{pred, end}}$) to be inaccurate while the foot is in the swing phase of the walking gait. Consequently, this renders the ALIP-based MPC footstep planner inadequate for agile and robust locomotion with multiple replanning steps during swing.    

Our goal is to circumvent this problem by appropriately designing an RL policy in conjunction with the MPC footstep planner, ensuring that the resulting policy incorporates the residual actions considering the full-order robot model. This approach overcomes the limitations of the ALIP-based MPC caused by modelling simplifications, without the need to solve a computationally challenging nonlinear optimization problem online with more complex dynamic models. Instead, it relies on simulation data collected offline that captures the robot's whole-body dynamics, allowing for efficient online deployment without substantial computation. 

\section{APPROACH}
\label{sec:solution_approach}

This section introduces our proposed approach to design a footstep policy that enables agile and robust bipedal locomotion by combining model-based optimal control and model-free RL methods. 

\subsection{System Overview}
Our method combines the ALIP-based MPC described in~\autoref{sec:alip_mpc} with an RL approach. The RL policy aims to impose supplementary actions onto the MPC to amplify the agility and robustness of the locomotion capability. 
Unlike the end-to-end RL, this synthesis provides sample-efficient and reliable policy learning by utilizing the MPC solution as prior knowledge. Furthermore, this fusion enhances the resulting policy's capability to adapt to challenging scenarios that the MPC process alone cannot address due to modelling simplifications.

The proposed learning framework, illustrated in Fig.~\ref{fig:rl_framework}, devises a residual footstep policy that generates task-space actions based on the robot's current states and gait commands. In each training episode, the gait commands are randomly sampled from a predefined range. Given a swing duration, the MPC process produces a dynamically consistent footstep plan that satisfies the proxy robot's kinematic and friction cone constraints in the ALIP model manifold. 
Starting from the MPC footstep policy, the RL agent effectively explores the robot's nonlinear whole-body dynamics manifold, which is facilitated by our proposed MDP design. Upon completing the training process, the optimal footstep location, formulated by combining the residual footstep policy with the ALIP-based MPC solution, is utilized for locomotion control.

\subsection{MDP formulation}
Our goal is to learn a residual policy $\pi_r$ to achieve an improved final policy $\pi$: 
\begin{align}
    \pi(s) = \pi_0(s) + \pi_r(s),
\end{align}
where $\pi_0(s)$ is the initial policy generated from the ALIP-based MPC process.
Given an MDP $\mathcal{M}$ process with a fixed initial policy $\pi_0(s)$, we define the residual MDP $\mathcal{M}^{(\pi_0)} = (\mathcal{S}, \mathcal{A}, \mathcal{P}^{(\pi_0)}, R^{(\pi_0)}, \gamma)$ similarly to the approach in~\cite{Silver2018ResidualLearning}, where:
\begin{equation*}
\begin{aligned}
P^{(\pi_0)}(s_k, a_k, s_{k+1}) &= P(s_k, \pi_0(s_k) + a_k, s_{k+1}), \\
R^{(\pi_0)}(s_k, a_k, s_{k+1}) &= R(s_k, \pi_0(s_k) + a_k, s_{k+1}).
\end{aligned}
\end{equation*}
The residual $\pi_r(s)$ is the policy in the residual MDP, which can then be solved using standard RL algorithms. In the following subsection, we describe the details of the residual MDP.

\subsubsection{Action Space}

Considering the initial footstep policy, $\pi_0(s)$, obtained from the ALIP-based MPC described in~\autoref{sec:alip_mpc} where the initial action is defined as $a_k^0 = (\mathbf{u}_{\textrm{fp},k}^{0}, \gamma_k^{0})$, we design the residual action as follows:
\begin{align} \label{eq:MDP_action}
    a_k = (\mathbf{u}_{\textrm{fp},k}^{\textrm{res}}, \gamma_k^{\textrm{res}}),
\end{align}
where $\mathbf{u}_{\textrm{fp},k}^{\textrm{res}} \in \mathbb{R}^2$ denotes the position deviation from the footstep position $\mathbf{u}_{\textrm{fp},k}^{0}$ in the x and y directions, and $\gamma_k^{\textrm{res}} \in \mathbb{R}$ represents the angle deviation from the footstep yaw $\gamma_k^0$. Note that we only consider a 3-dimensional action space assuming the terrain slope is constant. However, this approach can be extended to a more general action space that includes foot SE(3) representations. This choice of residual action space allows the final policy to learn using the full-order nonlinear dynamics of the bipedal robot, ulinke the use of the initial footstep policy, which relies on the simplified ALIP model.   


\subsubsection{Observation Space}

The observation space uses the states given by the ALIP model. We augment it by including additional states related to the robot's task space and a gait-related temporal variable. This state design choice enables the residual policy to better learn from the robot's full-order dynamics. These states are the following:
\begin{align} \label{eq:MDP_state}
     {s}_{k} = (\sigma_{k}, {T_r}_{k}, \boldsymbol\alpha_{k}, \boldsymbol\psi_{k},  \boldsymbol\beta_k, a^0_{k-1} + {a}_{k-1}),
\end{align}
where $\sigma_k \in \{-1, 1\}$ is an indicator of which foot is in stance, ${T_r}_k \in \mathbb{R}_+$ is the remaining foot swing time, $\boldsymbol\alpha_k = \left(x_c, y_c, z_c, L_x, L_y, L_z \right) \in \mathbb{R}^6$ is the state of the ALIP model with addition of the z components, $\boldsymbol\psi_k = \left(\boldsymbol\phi_{k}^{\textrm{torso}}, \boldsymbol\phi_{k}^{\textrm{sw}}, \boldsymbol\omega_{k}^{\textrm{torso}}\right) \in \mathbb{R}^9$ is the robot's orientation-related states with $\boldsymbol\phi_{k}^{\textrm{torso}}, \boldsymbol\phi_{k}^{\textrm{sw}}$ being the orientations of the torso and swing foot in Euler angle respectively, and $\boldsymbol\omega_{k}^{\textrm{torso}}$ being the angular velocity of the torso. $\bm{\beta}_k = \left(L_x^{\textrm{offset}}, L_{y}^{\textrm{des}}, \gamma^{\textrm{des}}\right) \in \mathbb{R}^3$ is the desired commands at the end of the step, $\left(a^0_{k-1} + a_{k-1}\right) \in \mathbb{R}^3$ is the policy's action taken in the previous step. Note that the linear velocity of the CoM is not included in the observation space as it is implicitly captured in the angular momentum about the contact point in $\bm{\alpha}_k$.

Including the most recent action from the previous step in the observation space is essential to ensure the smoothness of the swing foot trajectory. During footstep replanning while swinging, significant differences between two consecutive footstep solutions can lead to discontinuous swing foot trajectories, often rendering the robot's locomotion unstable and jerky. Therefore, incorporating the action from the previous step into the observation space allows it to be utilized in the reward function, promoting smoother and more stable locomotion.

\subsubsection{Rewards}

Given the set of early termination states $\mathcal{T}$ in an episode:
\begin{align}
    \mathcal{T} = \left\{s \in \mathcal{S} \mid    (z_{c}, L_{x,y}) \notin \left[z_{c}^{\textrm{min}}, z_c^{\textrm{max}}\right] \times \left[ L_{x,y}^{\textrm{min}}, L_{x,y}^{\textrm{max}}\right]\right\},
\end{align}
we design the reward function to ensure velocity tracking of the robot while realizing stable and robust locomotion. Additionally, the reward design enables the RL agent to explore beyond the simplified model (ALIP) state space by formulating the reward associated with the augmented observation and action space discussed in the previous subsection. The reward function is defined as follows:
\begin{align} \label{eq:MDP_reward}
    R(s_k, a_k, s_{k+1}) = 
    \begin{cases}
        0 & \text{if } s_{k+1} \in \mathcal{T}, \\
        r_{\textrm{sw}} & \text{if } s_{k+1} \notin \mathcal{T} \cap \sigma_{k+1} = \sigma_{k}, \\
        r_{\textrm{end}} & \text{otherwise}
    \end{cases}
\end{align}
with:
\begin{align*}
    r_{\textrm{end}} &= r_{a} + r_{L_x} + r_{L_y} + r_{\gamma} + r_{z_H} + r_{\phi}, \\
    r_{\textrm{sw}} &= \widetilde{r}_{L_{x}} + \widetilde{r}_{L_{y}} + r_{\pi}, 
\end{align*}
where the details for each term are described as follows:
\begin{align*}
        r_a &= \text{alive bonus (constant value)} \\ 
        r_{L_x}(s_{k+1}) &= \text{Ker}_{L_x}(e_{L_{x,k+1}}), \\
        r_{L_y}(s_{k+1}) &= \textrm{Ker}_{L_y}(e_{L_{y, k+1}}), \\
        r_{\gamma}(s_{k}, s_{k+1}) &= \textrm{Ker}_{\gamma}(\phi^{\textrm{torso}}_{ \textrm{yaw},k+1}), \\ 
      r_{z_H}(s_{k+1}) &= -w_{z_H} \abs{r_{k+1,z} - z_{H}}, \\
      r_{\phi}(s_{k+1}) &= - w_{\phi} \, \norm{(\phi_{\textrm{roll},k+1}^{\textrm{torso}}, \phi_{\textrm{pitch},k+1}^{\textrm{torso}})}_2^2, \\
        \widetilde{r}_{L_x}(s_{k+1}) &= \textrm{Ker}_{\widetilde{L}_x} (e_{\bar{L}_x,k+1}), \\
      \widetilde{r}_{L_y}(s_{k+1}) &= \textrm{Ker}_{\widetilde{L}_{y}}(e_{L_y,k+1}), \\
        r_{\pi}(s_k, s_{k+1}) &= \textrm{Ker}_{\pi}(a_{k}-a_{k-1}), \\
\end{align*}
with
\vspace{-0.10in}        
\begin{align*}
    e_{L_{\{x,y\}, k}} &= \left(L_{\{x,y\},k}-L_{\{x,y\},k}^{\textrm{des}}\right), \\
    e_{\bar{L}_{x, k+1}} &= \max\left(0, \abs{L_{x, k+1}- L_x^{\textrm{offset}}} - L_x^{\textrm{main}} \right), \\
    \textrm{Ker}_{v}(e) &= \omega_{v}\exp(- \left( e/\sigma_{v} \right)^{2}).
\end{align*}

We have designed the above reward terms in order to enhance the tracking of the dynamics obtained by the ALIP-MPC process shown in~\eqref{eq:discrete_dyn}. Specifically, $r_\textrm{sw}$ and $r_\textrm{end}$ are utilized during the intra-step and leg switching phases, respectively. The term $r_a$ encourages the robot to take more steps, which is particularly effective at the beginning of the training. The terms $r_{L_x}, r_{L_y}$ and $r_\gamma$ incentivize the robot to follow the desired gait commands. The terms $r_{z_H}$ and $r_\phi$ promote tracking the desired CoM height and torso pitch and roll, which in our case are kept constant. Similarly to the terms $r_{L_x}$ and $r_{L_y}$, $\widetilde{r}_{L_x}$ and $\widetilde{r}_{L_y}$ encourage the tracking of the desired $L_x$ and $L_y$ in the intra-step phase. Finally, the term $r_\pi$ encourages the final footstep policy to avoid excessive variations between the previous action and the current action.


\subsection{Learning and Policy Network}

Because we want to encourage the RL agent to explore the unknown residual dynamics between the full-order and ALIP models, we optimize our policy utilizing Proximal Policy Optimization (PPO)~\cite{Schulman2017ProximalAlgorithms}. The residual policy, $\pi_r$, is parametrized by the neural network which outputs the 3-dimensional residual footstep location.

During the training phase, in order for the RL agent to encourage exploration, we employ the residual action drawn from the Gaussian distribution:
\begin{align}
    a_k \sim \mathcal{N}(\mu_\theta, \sigma(r)),
\end{align}
where $\mu_\theta$ and $\sigma(r)$ denote the mean and covariance of residual actions. $\mu_\theta$ is parameterized by $\theta$ while $\sigma(r)$ is a scheduled parameter that depends on both initial conditions and the training procedure. Our policy is structured with a multi-layer perception architecture with two fully connected hidden layers of 64 units and the $\tanh$ activation function. The output layer is bounded by a scaling factor to constrain the maximum value. To effectively facilitate the RL agent to start exploring around the MPC policy, we initialize the last layer of the neural network to be zero, similar to~\cite{Silver2018ResidualLearning}.

\section{EXPERIMENTAL RESULTS}
\label{sec:experimental_results}
\begin{figure}[t!]
    \centering
    \includegraphics[width=\columnwidth]{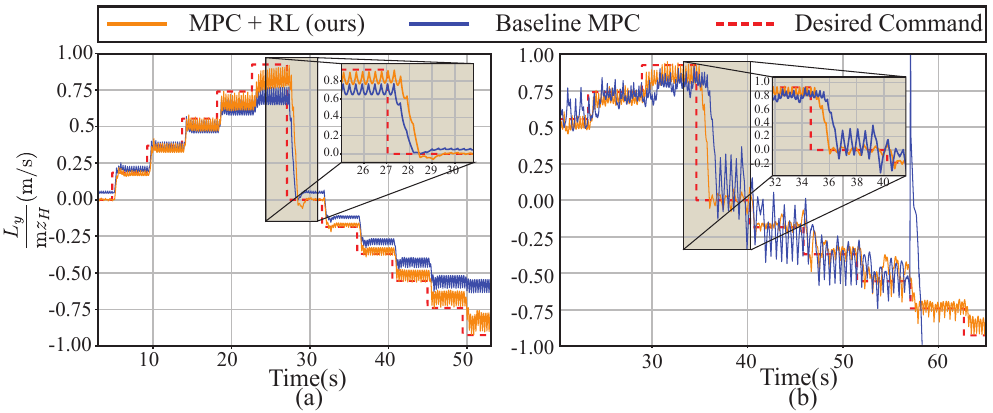}
    \caption{\textbf{Velocity tracking during forward walking:} Comparison of the baseline MPC and our proposed MPC + RL method: (a) Lower frequency policy  (b) Higher frequency policy}
    \label{fig:Ly_range_tracking}
\end{figure}
To evaluate the efficacy of our RL-augmented MPC footstep planner in achieving agile and robust omnidirectional locomotion as well as in traversing challenging terrains, we conducted several experiments in simulation. The baseline approach against which our proposed method was compared is the state-of-the-art MPC-based footstep planner in~\cite{Gibson2022Terrain-AdaptiveConstraints}. Additionally, to analyze the effects associated with the footstep planning frequency (i.e. the number of times we replan the trajectories during each footstep duration), we formulated a variation of our proposed MDP, that allows for planning footsteps once per foot swing, right at moment of the foot switching. This MDP is formulated by excluding intra-step dynamics-related components ($T_{r_k}$ and $a^0_{k-1} + a_{k-1}$ in~\eqref{eq:MDP_state} and $r_{\textrm{sw}}$ in~\eqref{eq:MDP_reward}). For more details on the experiments conducted in this section, please refer to the accompanying video.

\subsection{Experiment Setup}
We performed simulations on the 25-DoF Humanoid robot, DRACO 3~\cite{Bang2023ControlBody}, using Pybullet~\cite{Coumans2016PyBulletLearning}. All control modules employed in this paper are written in C++ and integrated with Python using Pybind11~\cite{WenzelJakobandJasonRhinelanderandDeanMoldovan2017Pybind11Python}, enabling their use during the simulations and RL policy training. Throughout all test cases, we maintained uniformity by adhering to the baseline MPC specifications, including foot swing height, MPC prediction horizon, and MPC weights, and by using WBC~\cite{Bang2023ControlBody} with identical task setups and gains. 

During training, to ensure that the data is collected at a consistent rate, we maintained a fixed update rate of 114~\si{Hz} for the high-frequency MPC and 5~\si{Hz} for the low-frequency MPC to generate the desired footholds. Meanwhile, the WBC process calculated joint commands at approximately 600~\si{Hz}. The two different RL policies operated at their respective MPC frequencies, labeled as the high-frequency (planning multiple times per step) and low-frequency (planning once per step) policies. These configurations expedited the training process without the need to solve the MPC process at every WBC control loop. 

\begin{figure}[t!]
    \centering
    \includegraphics[width=\columnwidth]{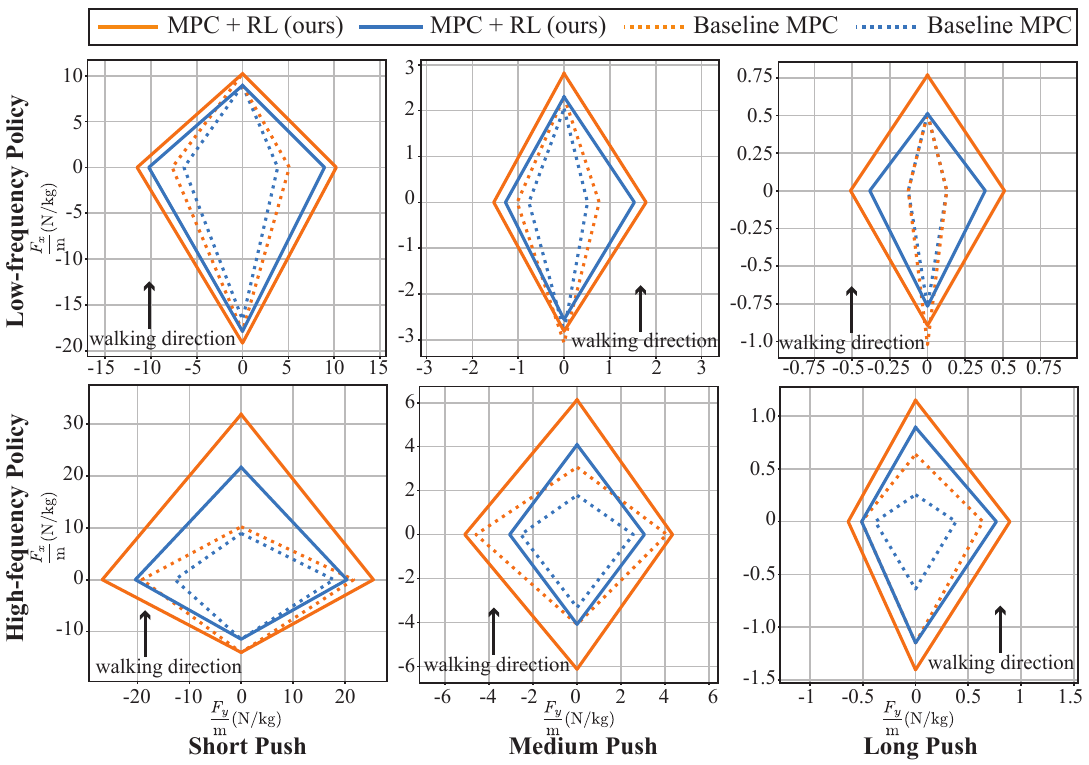}
    \caption{\textbf{Force perturbation tests:} At each evaluation episode, incremental perturbation forces (pushes) were applied until the robot lost balance. The blue curve represents the minimum force at which the robot failed to maintain balance. The orange curve represents the maximum force with which the robot consistently maintained balance when starting the evaluation episode with that force. Unlike the blue curve, independent of previous perturbations. (Top row): Low-frequency policy (Bottom row): High-frequency policy.}
    \label{fig:disturbance_diagram}
\end{figure}
\subsection{Simulation Results}
\subsubsection{Velocity Tracking}
This experiment aims to validate that the proposed approach can achieve fast, reliable walking in both sagittal and lateral directions. In the training setup, we randomized the velocity command between $\frac{L_y^{\textrm{des}}}{mz_H} \in [-0.925, 0.925]~\si{m/s}$. We tested our proposed MPC + RL policy for tracking a velocity profile in different walking directions. Fig.~\ref{fig:Ly_range_tracking} shows the velocity tracking performance of our proposed policy against the baseline MPC in the sagittal direction. Unlike the baseline MPC, our policy demonstrated excellent tracking of the gait commands while walking, both in steady-state and through transient commands. We note that training the RL agent for different velocity commands within an episode enabled effective tracking of aggressive changes in the velocity command, as demonstrated in Fig.~\ref{fig:Ly_range_tracking}(b). This performance is crucial for enabling the robot to walk at human speeds, rapidly accelerating and decelerating for practical tasks. Note that we did not train for aggressive changes in the policy results shown in Fig.~\ref{fig:Ly_range_tracking}(a). 

For lateral walking, we used a training setup similar to the one used for sagittal walking, but with $L_x^{\textrm{des}}$ as the target angular momentum component. Our combined MPC+RL policy efficiently enabled the robot to track velocities within the range of $[-0.83, 0.83]~\si{m/s}$ while avoiding self-collisions. In contrast, the baseline MPC was only able to track lateral velocities within the range of $[-0.6, 0.6]~\si{m/s}$, as indicated by the accompanying video.

\begin{figure}[t!]
    \centering
    \includegraphics[width=\columnwidth]{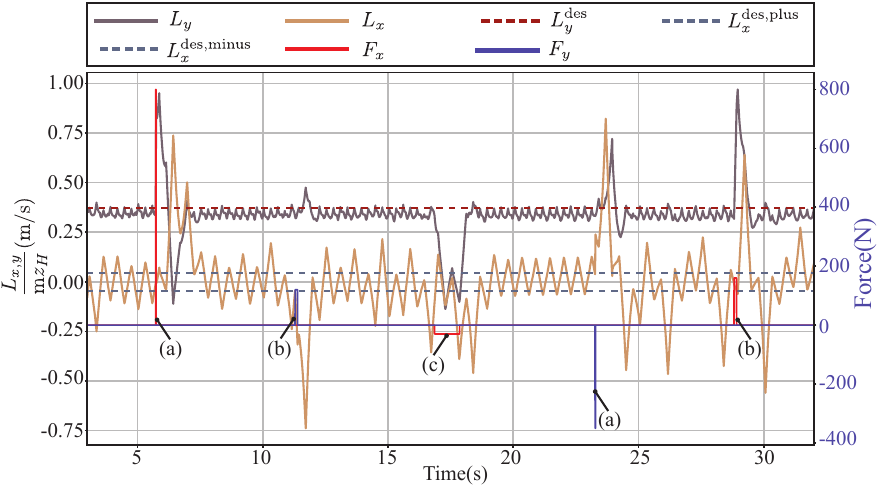}
    \caption{\textbf{Velocity tracking under external disturbance:} The performance of the proposed high-frequency policy under various types of pushes: (a) Short push, (b) Medium push, (c) Long push.}
    \label{fig:disturbance_tracking}
\end{figure}
\subsubsection{Turning in place}
To evaluate the performance of our approach on tasks where the ALIP-based MPC process fails, we consider a continuously turning task. The difficulty of such motion arises from the use of the ALIP model, which does not include rotational dynamics. During training, we considered the desired turning rate $\dot{\Theta}_z^{\textrm{des}} = 1.27~\si{rad/s}$ corresponding to $\gamma^{\textrm{des}} = 0.35 ~\si{rad/step}$. 

For such scenario, we define the performance metric of the policy by its capability to perform a full $360^{\circ}$ turn without losing balance while tracking the desired yaw rate commands. The high-frequency policy enabled the robot to accomplish an average number of $2.7$ turning motions per episode with a yaw rate of $1.27$~\si{rad/s}, which was $350\%$ better than the baseline MPC policy, which consistently failed to achieve one full $360^{\circ}$ turn. On the other hand, the low-frequency policy allowed for tracking a yaw rate of up to $1.9$~\si{rad/s} without falling, while the baseline MPC was only able to perform an average of 7 steps per episode corresponding to $40\%$ completion of a full turn. This improved stability while turning-in-place is achieved thanks to our MPC + RL footstep policy, effectively exploring beyond the base footstep solution space limited by the use of the simplified ALIP model.


\subsubsection{Robustness against Disturbances}
In this experiment, we study the effectiveness of our policies in handling external disturbances. During training, 
we considered two types of force disturbances (pushes): long push with a duration of 1~\si{s} and short push with a duration of 0.0175~\si{s}. We selected 35~\si{N} for the long push and 350~\si{N} for the short push, where the magnitude of the force is close to the robot's weight. We randomly sampled the type of pushes, as well as their direction (either sagittal or lateral) and timing (any time during the foot swing) applied to the robot's torso link. We chose a forward walking gait at 0.37~\si{m/s}.

We assessed the robustness of our policy in rejecting external disturbances by applying three types of pushes, including the two types of pushes during training and a medium push with a duration of 0.1~\si{s}. Fig.~\ref{fig:disturbance_diagram} shows the maximum force magnitude (scaled by the robot's mass) for different directions that our policies and the baseline MPC can withstand without losing balance. Our policies outperform the baseline MPC in resisting pushing forces for all directions and all types of pushes while maintaining the forward walking motion. Notably, our higher frequency policy demonstrated superior performance to the lower frequency policy while showcasing remarkable robustness to the extensive range of perturbation not considered during training. This result can be attributed to our novel MDP formulation to recompute new footstep locations during the swing motion. Moreover, the higher frequency policy excelled in tracking velocity effectively under various intensive external pushing forces, as depicted in Fig.~\ref{fig:disturbance_tracking}.

\begin{figure}[t!]
    \centering
    \includegraphics[width=\columnwidth]{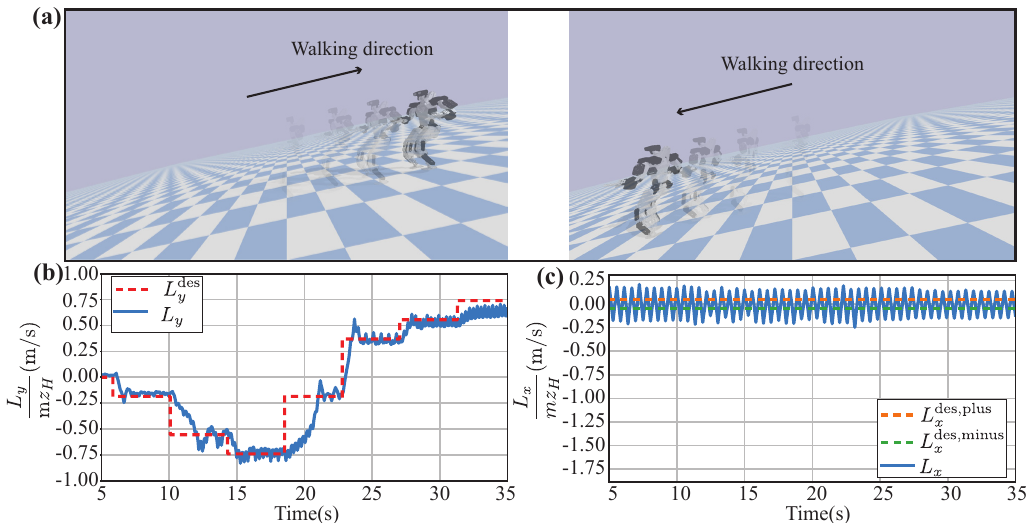}
    \caption{\textbf{Sloppy terrain walking:} The tracking performance of the proposed high-frequency policy on a 11.5-degree slope: (a) Snapshots of walking in the sagittal direction, (b) Velocity tracking performance in the sagittal direction, (c) Velocity tracking performance in the lateral direction.}
    \label{fig:slope_walking}
\end{figure}
\subsubsection{Walking in Sloppy Terrain}
To assess the applicability of our policy to challenging terrains, we tested it in terrains with slopes up to $11.5$ degrees by evaluating the speed tracking performance. In this scenario, we assumed the robot's perception system accurately estimated the slope of the terrain. In contrast to the baseline MPC, our proposed high-frequency policy demonstrated a good speed-tracking performance while walking in a sagittal direction on every slope. Fig.~\ref{fig:slope_walking} shows a tracking performance of our high-frequency policy on a slope of $11.5$ degrees.

\subsubsection{Sample Efficiency}
To validate the sample efficiency of our approach, we compared it with an RL-only approach that learns the task space policy similarly to~\cite{Castillo2023TemplateLocomotion}, but with modifications to allow re-planning during the foot swing duration. Our approach demonstrated a faster convergence rate 
thanks to the initial MPC policy.

\section{CONCLUSION AND FUTURE WORK}
\label{sec:conclusion}
This work presents a unique online bipedal footstep planning framework capable of reliably and efficiently adjusting planned footsteps to achieve precise and robust velocity tracking for agile and robust dynamic locomotion. 
Our approach combines a simplified model-based MPC footstep policy with a model-free RL policy. By leveraging RL's flexibility and adaptability, the resulting policy efficiently handles more complex locomotion tasks that challenge the simplified model-based MPC footstep planning process. We showcase our experimental results during various locomotion tasks, including tracking a wide range of walking speeds while traversing flat and sloppy terrain and exhibiting reliable turning behaviors. This demonstrates a significant enhancement in dynamic locomotion performance compared to the baseline simplified model-based MPC.  

While our approach has shown effectiveness in dynamic locomotion, there are exciting areas for future exploration. First, the action space of our policy is limited to a 3-dimensional footstep task space. This action space can be extended to a larger dimensional task space (e.g., by including CoM height and orientation) to further handle modelling errors introduced by the use of simplified models, leading to improved stability and adaptability of locomotion. Second, as an extension of our work, loco-manipulation policies can be similarly constructed by integrating the existing model-based (e.g., centroidal dynamics~\cite{Murooka2022CentroidalMotion}) controller with an RL policy to enhance the robustness and versatility of the model-based controller.


\addtolength{\textheight}{-12cm}   





\section*{ACKNOWLEDGMENT}
This work was supported by the Office of Naval Research (ONR), Award No. N00014-22-1-2204.


\bibliographystyle{IEEEtran}
\bibliography{references}

\end{document}